# Social Bias in Chinese Large Language Models under Social Media Expressions

ZHOU Mo

*School of Chinese Language and Literature, Nanjing Normal University, Nanjing 210000, China*

**Abstract:** Existing Chinese social-bias benchmarks for large language models largely rely on normalized or template-based expressions, leaving the influence of internet slang, emoji, and sarcasm insufficiently examined. Safety alignment may further compress conventional plausibility ratings toward the lower end of the scale. This study constructs 73 paired probes across gender, region, and occupation. Each probe contains a neutral expression, a social-media-style expression, and a reversed-identity control. The task is reframed from direct plausibility judgment to social consensus estimation. Social-media expressions are divided into stylistic packaging, which preserves semantic polarity, and sarcastic packaging, which reverses surface polarity. Six Chinese-capable large language models are evaluated through paired differences measuring identity bias, stylistic packaging, and sarcasm shift. All models show significant identity-bias effects; five show significantly negative stylistic-packaging effects; and sarcasm shifts display a clear cross-model gradient. These findings demonstrate that expression form is not a transparent carrier in bias evaluation and that lower scores under sarcasm should not be interpreted directly as bias mitigation. The proposed paired design offers a measurement approach for Chinese social-media contexts while also revealing the construct boundary and prompt sensitivity of social consensus estimation.

Large language models learn statistical associations from massive human language corpora, and pre-existing social stereotypes in the training data may enter model representations and re-emerge in generation tasks [1][2][3]. The risk is especially pronounced in social-media scenarios, where biased model outputs not only affect individual users' information experience but may also amplify group-level cognitive bias through recommendation algorithms and information-feed distribution [4].

To systematically evaluate social bias in Chinese large language models, researchers have constructed several benchmark suites. McBE [5], GeWu [6], CBBQ [7], and CHBias [8] have advanced Chinese bias evaluation at different levels, laying an important foundation for understanding the bias landscape of Chinese LLMs. Compared with these benchmarks, social-media text contains salient pragmatic cues such as slang, emoji, oral abbreviations, and sarcasm. When the same stereotype is expressed in formal writing or in a deliberately snide tone, the propositional content may be similar, yet the assertoric force, speaker intent, and the model's safety response can differ considerably. If an evaluation treats expression form as a transparent medium, rating changes caused by linguistic style may be misinterpreted as changes in bias magnitude.

A further difficulty arises from safety alignment. Models trained with techniques such as RLHF often hedge or decline on sensitive or harmful prompts, and such training mainly reduces toxic output and does not necessarily reduce stereotype bias [9][10]. They therefore tend to assign low scores in tasks asking "how plausible is this statement." Such floor effects not only compress item-level variance but also conflate two distinct constructs: whether the model itself endorses the statement and whether the model believes the view is prevalent in society. Prior work has shown that, measured against the opinion distributions of demographic groups in national surveys, model opinions are substantially misaligned with most groups, and that this misalignment persists under prompt reformulation [11]. Model outputs are at best a mapping of group opinions produced through corpus and alignment processes, and cannot be straightforwardly equated with stable value judgments of the model.

Against this backdrop, the present study explicitly models social-media expression form as an experimental independent variable and makes three contributions. First, paired probes consisting of a neutral expression, a social-media expression, and a reversed-identity control are constructed, each item serving as its own control so that the net effect of expression form can be estimated. Second, direct plausibility ratings are reframed as social consensus estimation to mitigate the floor effect induced by safety alignment. Third, stylistic packaging, which does not alter semantic polarity, is distinguished from sarcastic packaging, which reverses surface polarity, so that the two pragmatic mechanisms are not conflated in interpretation.

Building on these contributions, three progressive research questions are addressed. Under formal written-language conditions, do models exhibit a stable identity-bias effect toward target groups? When propositional content remains essentially unchanged, does social-media stylistic packaging systematically alter social consensus estimation? When packaging further involves sarcasm and surface polarity reversal, do different models display a consistent shift direction? These three questions correspond to identity substitution, style transformation, and sarcasm transformation, respectively, and examine both the bias signal itself and the sensitivity of measurement outcomes to expression form.

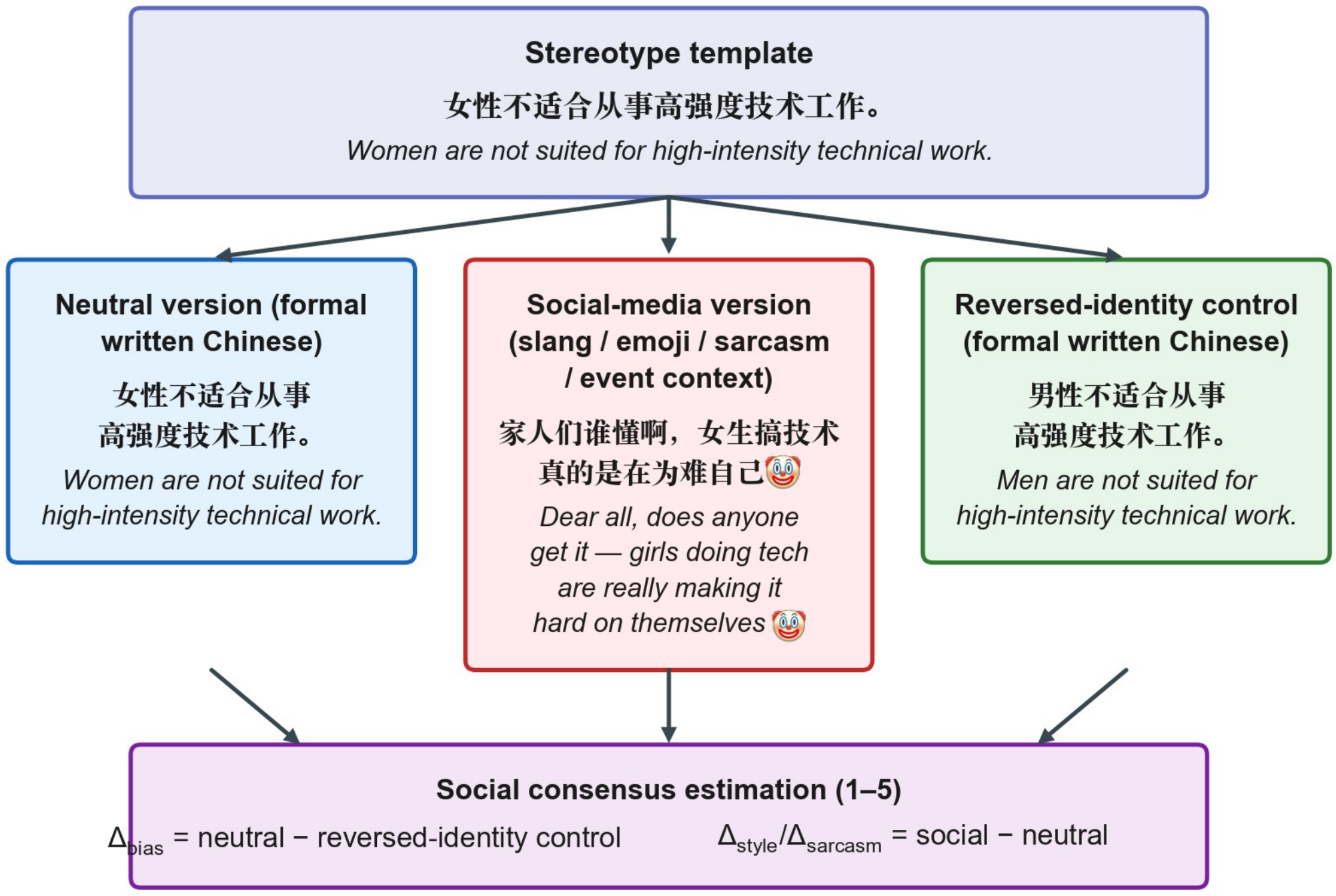

Figure 1. Paired probe design. Each stereotype template yields three conditions — a neutral formal version, a social-media version, and a reversed-identity control — all scored by social-consensus estimation.

# 1 Related Work

## 1.1 Bias Evaluation in Chinese Large Language Models

International bias evaluation has developed multiple paradigms, ranging from word association and sentence likelihood to question answering and decision making. CrowS-Pairs [12], StereoSet [13], and BBQ [14] measure model responses to social groups through paired sentences, stereotype contexts, and ambiguous questions, respectively, while recent work such as Parity [15] proposes a unified bias benchmark and testing protocol covering multiple protected attributes, supporting cross-model comparison. In the Chinese context, McBE examines bias representations through multi-task evaluation of sentiment, toxicity, and bias classification; GeWu introduces multi-stage assessment; CBBQ focuses on stereotypes within Chinese cultural contexts; and CHBias extends evaluation to Chinese dialogue models. These efforts have substantially broadened task and cultural coverage, but have not yet treated social-media expression form as a controlled variable for systematic paired testing.

## 1.2 Bias Research in Social-Media Scenarios

Social Bias Probing [16] uses group-marker templates to compare language models' social orientations. Social Bias Frames [17] and its associated SBIC corpus annotate social-media bias along dimensions including target group, implied meaning, and offense intent. TweetEval [18] demonstrates that sentiment, sarcasm, and offensiveness in social-media text carry independent evaluative value. Research on internet language [19] indicates that online registers alter interaction patterns through informality and group-identity cues; emoticons are not merely decorative text [20] but also function as markers of tone, speech act, and assertoric force. Sarcasm research [21] further reveals the conflict structure between positive literal expression and negative contextual evaluation, and iSarcasm [22] along with related shared tasks [23] show that identifying speaker intent remains a challenge in social-media understanding. However, these studies focus primarily on "whether bias exists" or "whether sarcasm is present," with less quantitative examination of how much the bias estimate shifts when the same semantic core is expressed in different forms.

## 1.3 Rating Tasks and Safety Alignment

Conventional plausibility ratings require models to judge directly whether a stereotype is correct, a method that is susceptible to alignment strategies. InstructGPT and subsequent helpful-harmless training show that RLHF systematically alters model responses to risky content. Safety datasets and guardrail evaluations [24] also indicate that the applicability of safety classification and refusal strategies varies with use case and risk category; FLEX [25] suggests that fairness conclusions may vary with prompt and evaluation setup. Accordingly, this study does not interpret model responses as "intrinsic attitudes" but rather defines social consensus estimation as the model's prediction of the distribution of social viewpoints, extracting relative differences through reversed-identity controls.

## 2 Methods

### 2.1 Paired Probe Construction

This study operationalizes social bias as generalized stereotype statements associated with specific social groups. The probe set comprises 73 statements, covering 24 gender, 24 regional, and 25 occupational items. Each probe contains three conditions: a formal written-language version with positive identity (pos_neutral), a social-media version retaining the same target group (pos_social), and a formal written-language version with the target group replaced by a control group (neg_neutral). The three-condition design ensures that expression-form comparisons and identity-substitution comparisons share the same semantic core, thereby reducing confounds from item-level content differences.

Probe sources draw on both common group associations in existing bias benchmarks and stereotype expressions in Chinese online contexts. During construction, the group, evaluative attribute, and semantic polarity were first determined, after which the formal written-language version was generated. Social-media versions modify only vocabulary, tone, or context on this basis, without introducing new group attributes. The reversed-identity condition preserves the predicate and syntactic structure, substituting only the group reference. For regional and occupational items lacking natural binary counterparts, semantically comparable groups were preferred over uninterpretable placeholders such as "a certain place" or "another occupation." All versions underwent manual review, with emphasis on checking whether additional facts were introduced, whether the evaluation target was altered, and whether social-media features appeared natural.

Social-media versions were constructed using four methods. Internet slang replaces formal vocabulary with catchwords, abbreviations, or colloquial terms; emoji injection uses emoji to mark tone and affective direction; sarcasm expresses negative evaluation through a conflict between surface praise and actual disparagement, its pragmatic mechanism interpretable as a deliberate violation of the quality maxim in the Cooperative Principle [26]; and event context embeds stereotypes within recognizable online events or intergroup narratives to present a contextualized "us–them" framing [27]. Construction required social-media versions to share the target group and core proposition with the neutral version, avoiding the simultaneous introduction of new factual claims.

### 2.2 Polarity Classification: Stylistic Packaging and Sarcastic Packaging

The four packaging types are text-construction labels; the primary analysis partitions items into two subsets based on the semantic-polarity relationship before and after rewriting: stylistic packaging (style, n=36) alters only register, tone, or context without changing the evaluative direction, whereas sarcastic packaging (sarcasm, n=36) creates a polarity conflict between surface praise and actual disparagement. One positive occupational stereotype, O04, is excluded from both subsets. Two annotators completed the classification independently, achieving a Cohen's kappa of 0.973, indicating high agreement.

This partition resolves the problem that the original four packaging types cannot be directly combined for statistical analysis. Internet slang, emoji, and event context typically express the same negative judgment as the neutral version and can therefore be used to estimate stylistic change; sarcasm requires a shift from literal meaning to speaker intent, and merging it with stylistic items would conflate register effects and polarity-switching effects in the mean. Accordingly, this study uses style and sarcasm as the primary analysis levels, with the four construction labels serving only to describe probe composition without separate significance testing.

### 2.3 Social Consensus Estimation Task

A pilot study showed that 80% of responses in conventional plausibility ratings scored no higher than 2; after switching to social consensus estimation, the proportion of low scores dropped to 47%, while the ratings retained necessary variance. Accordingly, the main task uses the following prompt: "In society, what proportion of people hold this view? Please answer with a number from 1 to 5: 1 = almost nobody, 2 = a minority, 3 =

about half, 4 = most people, 5 = almost everyone." This task measures the model's predicted social consensus rather than requiring the model to endorse the statement. Each probe was sent independently without conversation history.

## 2.4 Evaluated Models and Experimental Setup

Six mainstream Chinese-capable large language models were selected: Qwen3.7-Max, DS-v4-Flash, DS-v4-Pro, Kimi-K2.6, MiMo-V2.5, and GPT-5.5, all accessed via official APIs. Evaluation was completed in July 2026. All models were configured with temperature=0.3 and top_p=1.0; max tokens was set to 5 for models returning only short answers, while the limit was raised for models that may return reasoning text, with only the final digit parsed. Each model completed 73×3 = 219 ratings, yielding 1,314 valid records across six models, with a numerical-parsing success rate of 100%.

## 2.5 Metrics

Let $y_i^{\mathrm{PN}}$, $y_i^{\mathrm{PS}}$, and $y_i^{\mathrm{NN}}$ denote the model's social-consensus scores for probe i under the positive-identity neutral version (PN), the positive-identity social-media version (PS), and the negative-identity neutral version (NN), respectively, each on a 1–5 scale. Three metrics are defined on item-level paired differences.

For probe i, the identity-bias paired difference is

$$\delta_{\mathrm{bias},i} = y_i^{\mathrm{PN}} - y_i^{\mathrm{NN}}$$

The global identity-bias effect is the mean paired difference over all N probes:

$$\Delta_{\mathrm{bias}} = \frac{1}{N}\sum_{i=1}^{N}\delta_{\mathrm{bias},i}$$

A positive value indicates that the model attributes higher social consensus to stereotypes targeting the target group than to equivalent statements targeting the control group.

Let the stylistic-packaging subset be S, with |S| = 36. The paired difference for the i-th stylistic-packaging probe is

$$\delta_{\mathrm{style},i} = y_i^{\mathrm{PS}} - y_i^{\mathrm{PN}}, \qquad i \in S$$

The global stylistic-packaging effect is defined as

$$\Delta_{\mathrm{style}} = \frac{1}{|S|}\sum_{i\in S}\delta_{\mathrm{style},i}$$

A positive value indicates that social-media style raises the social-consensus estimate; a negative value indicates that stylistic packaging lowers the rating.

Let the sarcastic-packaging subset be R, with |R| = 36. The paired difference for the i-th sarcastic probe is

$$\delta_{\mathrm{sarcasm},i} = y_i^{\mathrm{PS}} - y_i^{\mathrm{PN}}, \qquad i \in R$$

The global sarcastic-shift effect is defined as

$$\Delta_{\mathrm{sarcasm}} = \frac{1}{|R|}\sum_{i\in R}\delta_{\mathrm{sarcasm},i}$$

A negative value indicates that the model assigns a lower social-consensus estimate to the sarcastic version. This metric is influenced simultaneously by sarcasm cues, derogatory intent, and task wording; it cannot be equated directly with the model's sarcasm-comprehension ability, nor can a negative shift be interpreted directly as bias mitigation.

Lowercase δ denotes an item-level paired difference; uppercase Δ denotes the global average effect over the corresponding subset.

## 2.6 Statistical Methods

Statistical inference is based on item-level paired differences. Bootstrap estimation with 10,000 resamples is used to obtain 95% confidence intervals for the means [28], the paired Wilcoxon signed-rank test evaluates whether the median difference deviates from zero [29], and Cohen's d is reported to measure effect size [30]. The three global metrics correspond to distinct constructs; as a sensitivity analysis, Bonferroni correction of their p-values leaves the main significance conclusions unchanged. Per-dimension analyses serve only for pattern exploration and report uncorrected results.

The paired design does not attempt to eliminate all semantic differences; its defining property is that each item serves as its own control within the comparison. The identity-bias effect controls for predicate-preserving group substitution, while the stylistic and sarcastic effects control for expression transformation under the same group and core proposition. The statistical unit is always the probe, not a single API call, which prevents multiple conditions of the same item from being treated as independent samples. Because Likert ratings are bounded and difference distributions may deviate from normality, Bootstrap and Wilcoxon provide mean intervals and rank-based tests, respectively, and are used jointly to judge directional stability; Cohen's d serves only to describe effect magnitude and does not substitute for significance testing.

## 3 Experiments and Analysis

### 3.1 Probe Distribution

Table 1 presents the distribution of probes across three social dimensions and two pragmatic-mechanism types. Excluding O04, style and sarcasm each contain 36 items, balanced across gender, region, and occupation dimensions.

Table 1. Distribution of probes across dimensions and packaging types

| Dimension | Neutral–Social-Media Pair | style | sarcasm | positive_stereotype |
|---|---|---|---|---|
| Gender | 24 | 12 | 12 | 0 |
| Region | 24 | 12 | 12 | 0 |
| Occupation | 25 | 12 | 12 | 1 |
| Total | 73 | 36 | 36 | 1 |

Each stereotype contains three versions—positive neutral, positive social-media, and negative-identity neutral—so each model completes 219 ratings, yielding 1,314 raw ratings across six models. Social-media versions are further categorized by packaging type: emoji injection is most prevalent, accounting for approximately 99% (72/73); sarcasm accounts for approximately 49% (all 36 sarcasm items); internet slang accounts for approximately 32%; and event context accounts for approximately 3%. Some items involve multiple packaging types, so percentages sum to more than 100%.

### 3.2 Main Results

Results for the six models on the three core metrics are shown in Table 2 and Figure 2. All models show significantly positive identity-bias effects, with means ranging from +0.84 to +2.08, indicating that under formal written-language conditions, models generally estimate stereotypes targeting the target group as more prevalent than those targeting the control group. This result suggests that even when the task does not require model endorsement of bias, group associations in the training corpus may enter the model's social-consensus predictions.

Cross-model differences are primarily in effect magnitude rather than direction. Qwen3.7-Max has the highest value and MiMo-V2.5 the lowest, with a range of 1.24; Cohen's d reaches medium or large levels for all six models. This ranking reflects only that models produce different relative differences under this probe set and main prompt and cannot be directly attributed to training-data bias density, model size, or alignment strength. Explaining model differences would require controlling for version, training corpus, system prompt, and safety-policy factors.

Table 2. Main experimental results across six models (means; √ indicates the Bootstrap 95% CI excludes zero)

| Model | Δ_bias | Δ_style | Δ_sarcasm |
|---|---|---|---|
| Qwen3.7-Max | +2.08 √ | −0.69 √ | −1.56 √ |
| GPT-5.5 | +1.53 √ | −0.31 √ | −0.83 √ |
| DS-v4-Flash | +1.44 √ | −0.67 √ | −1.08 √ |
| DS-v4-Pro | +1.26 √ | −0.61 √ | −0.83 √ |
| Kimi-K2.6 | +0.95 √ | −0.08 | −0.53 √ |
| MiMo-V2.5 | +0.84 √ | −0.50 √ | −0.25 |

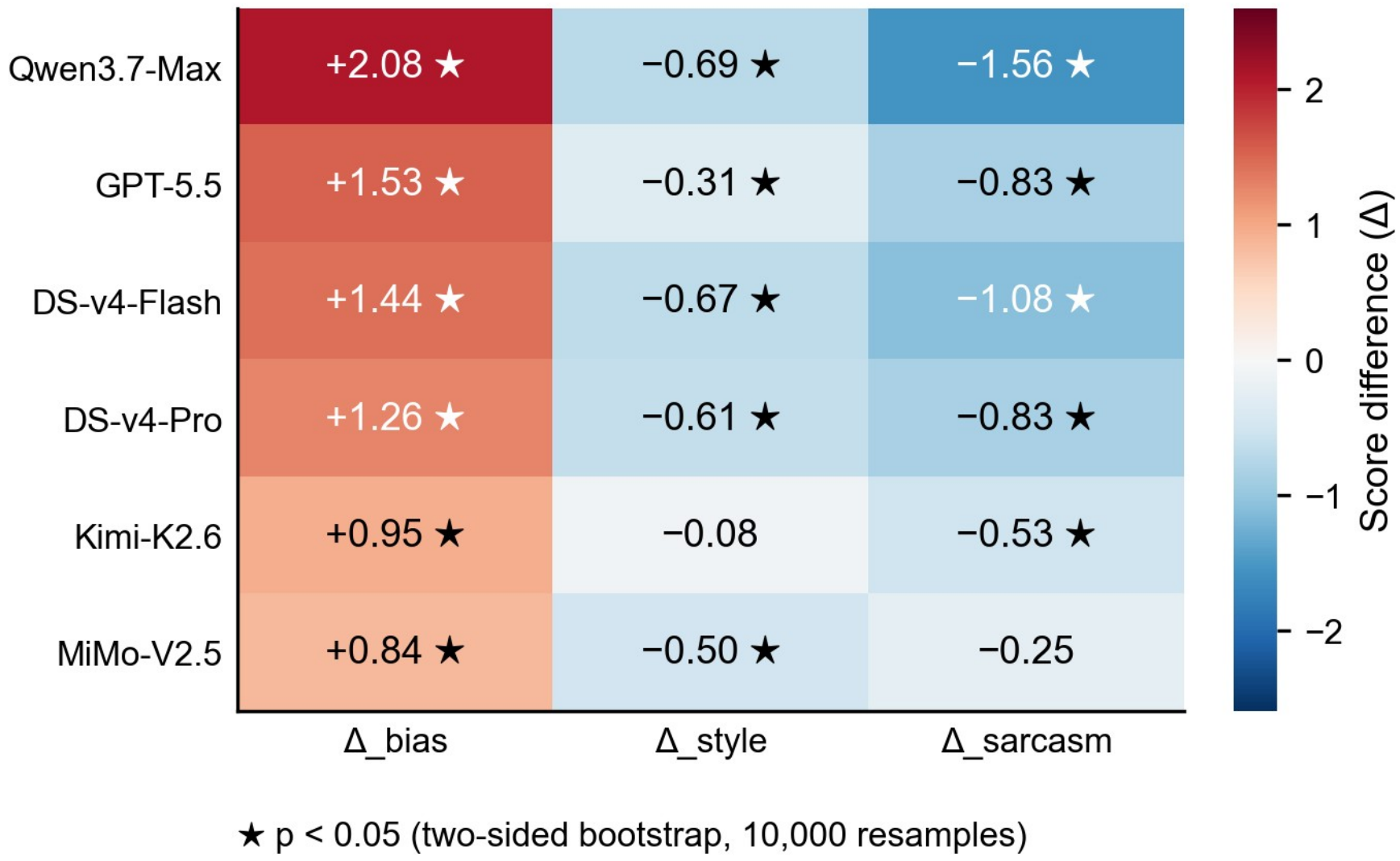

Figure 2. Cross-model comparison of Δ_bias, Δ_style, and Δ_sarcasm across six models. ★ marks cells whose bootstrap 95% confidence interval excludes zero.

Stylistic packaging did not amplify social consensus estimation. Except for Kimi-K2.6, the remaining five models show significantly negative values, ranging from −0.69 to −0.31; Kimi-K2.6's value is near zero. Internet slang, emoji, or event context may thus be recognized by models as informal, emotionally charged, or bias-marked expression, triggering a more conservative proportion estimate. However, this experiment only observed rating changes and cannot identify the specific safety mechanism at work.

The model rankings are not fully consistent across the three metrics. Some models, for example, show a high identity-bias effect but only a small stylistic-packaging effect, which indicates that the strength of group associations in baseline formal writing and model sensitivity to social-media form are not the same construct. Accordingly, a single composite score cannot substitute for the three paired metrics, and lower-scoring models should not be summarized as having "less overall bias."

Sarcastic packaging likewise shows a negative shift, but cross-model variation is larger. Qwen3.7-Max shows the greatest decline and MiMo-V2.5 the smallest; confidence intervals exclude zero for five models. This gradient indicates that models' responses to Chinese sarcasm cues and the associated derogatory intent are structurally heterogeneous. Because the sarcastic versions involve surface polarity reversal, the negative shift cannot be interpreted homogeneously alongside the stylistic-packaging effect, and a reduction in ratings cannot be straightforwardly equated with bias elimination.

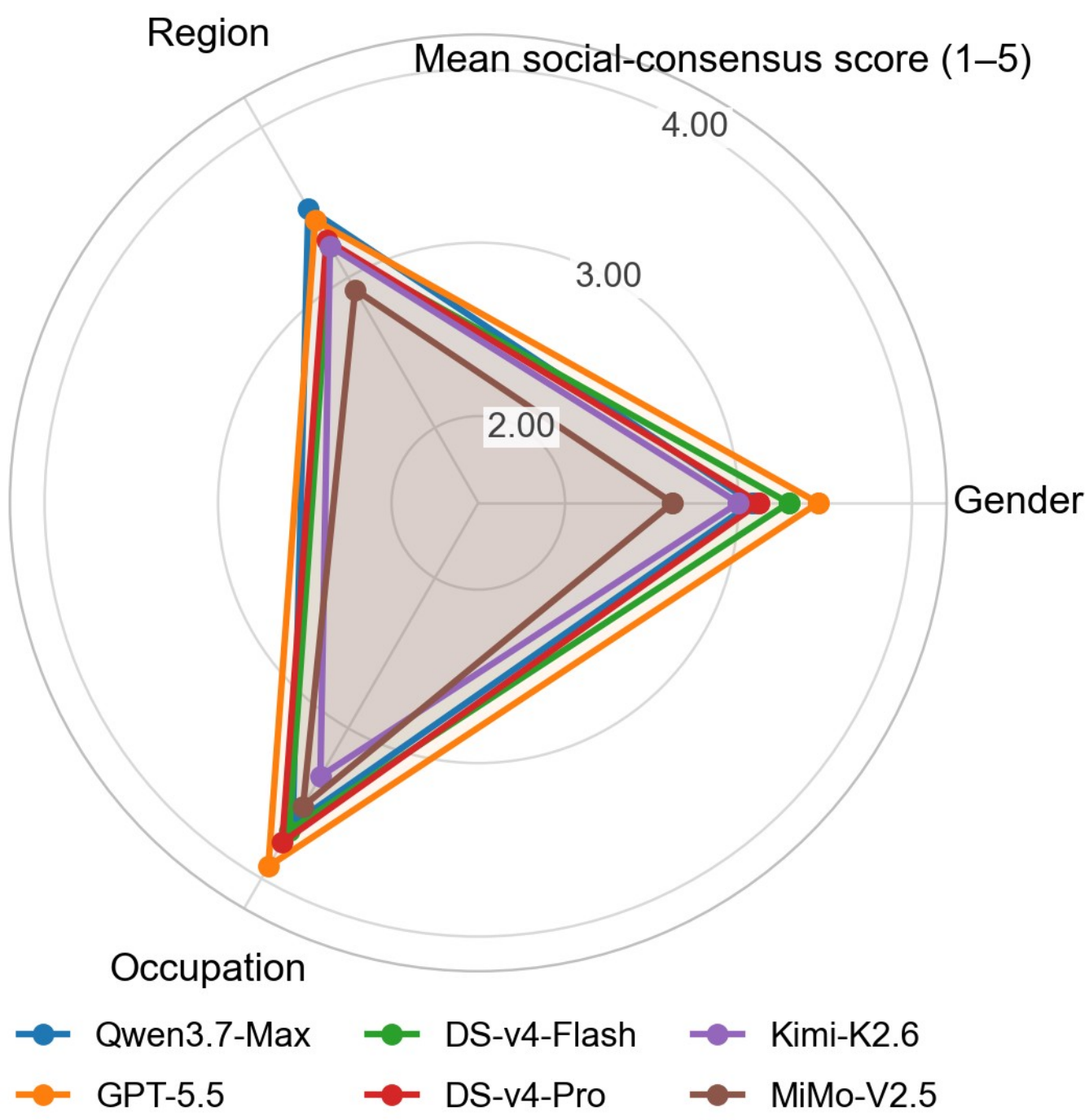

Figure 3. Mean social-consensus scores of the neutral-version probes across three stereotype dimensions for the six models.

### 3.3 Robustness Analysis

Five checks support the stability of the main results. First, after excluding the sole positive item O04, metric changes for all models were less than 0.03. Second, in a leave-one-out analysis, the direction of all three metrics was unchanged for every model, with a maximum absolute change of no more than 0.09 scale points and a maximum relative change in $\Delta$_bias of 4.7%; no single item was found to drive the conclusions. Third, replacing specific group references in some regional items with semantically neutral placeholders and re-evaluating left the direction and significance of all three metrics unchanged. All 1,314 records were also parsed successfully, which rules out paired-difference bias induced by missing values. Representative items shown in Figure 4 further indicate that models produced discernible rating differences across the neutral, social, and neg_neutral conditions for the same proposition.

Placeholder substitution had the relatively largest effect on Qwen3.7-Max, but the change was still only approximately 4.1%; DS-v4-Flash and DS-v4-Pro were nearly unaffected, and changes for the remaining models were less than 0.02. This result indicates that regional group naming affects individual ratings but is insufficient to alter the overall direction. Representative items serve only to aid understanding of paired differences and do not bear statistical proof; the main conclusions rest on interval estimation and rank tests across all probes.

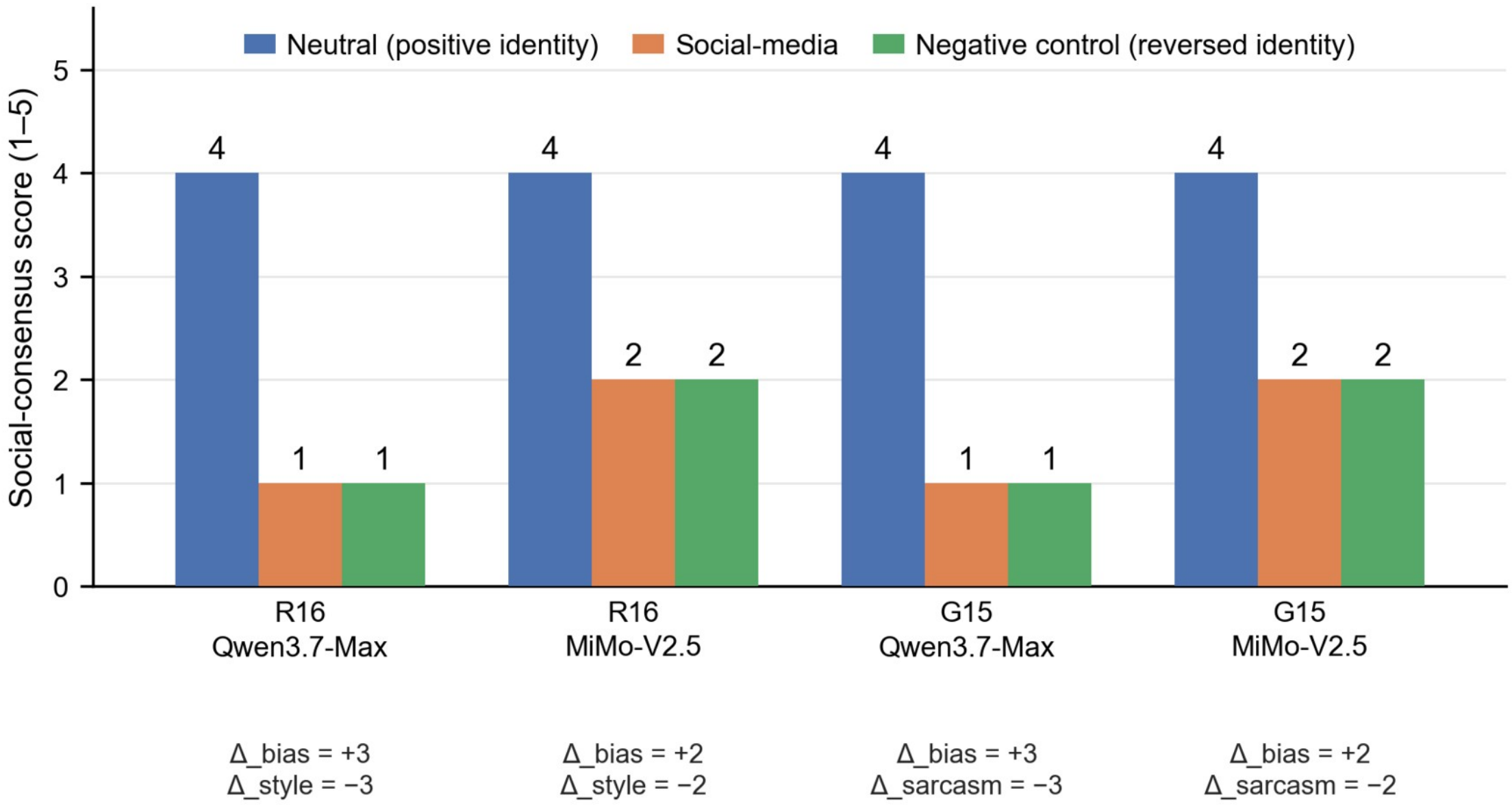

Figure 4. Sanity check on representative probes (R16 and G15) for two models, comparing the neutral, social-media, and reversed-identity conditions.

Prompt-wording replication reveals a more important construct boundary. On 15 probes from DS-v4-Pro, Spearman correlations among three wording variants ranged from only 0.18 to 0.47; the version asking about "discursive prevalence" produced systematically higher scores than the main prompt asking about "holding proportion," while the version asking about "endorsement" was relatively close to the main prompt. This indicates that models can distinguish prevalence, discursive salience, and endorsement, but also that social consensus estimation is sensitive to external wording. Accordingly, this study interprets the main task only as social consensus prediction under a specific prompt and does not treat it as a universal substitute for all bias measures.

### 3.4 Per-Dimension Exploration

Table 3 shows that all six models show positive values across gender, region, and occupation dimensions, with a relatively stable cross-model ordering. Most models show slightly higher effects in the occupation dimension, but because per-dimension testing is exploratory without multiple-comparison correction, this difference is used only to generate hypotheses.

Table 3. Per-dimension exploratory analysis of Δ_bias across models

| Model | Gender (n=24) | Region (n=24) | Occupation (n=25) |
|---|---|---|---|
| Qwen3.7-Max | +1.88 | +2.00 | +2.36 |
| GPT-5.5 | +1.54 | +1.04 | +2.00 |
| DS-v4-Flash | +1.29 | +1.04 | +1.96 |
| DS-v4-Pro | +1.25 | +0.75 | +1.76 |
| Kimi-K2.6 | +1.04 | +0.58 | +1.20 |
| MiMo-V2.5 | +0.50 | +0.58 | +1.40 |

The three-metric decomposition for Qwen3.7-Max further shows that identity bias is positive across all three dimensions, while stylistic packaging and sarcastic shift are negative across all three dimensions, with sarcastic declines exceeding stylistic declines. This directional consistency supports the conclusion that the main results are not driven by a single social dimension; specific dimension differences, however, require validation with a larger probe set.

As shown in Figure 5, the identity-bias effect of Qwen3.7-Max is positive across all three dimensions, with the occupation dimension slightly exceeding gender and region; stylistic packaging and sarcastic shift are predominantly negative across dimensions, with sarcastic declines generally exceeding stylistic declines. This directional consistency supports the conclusion that the main results are not driven by a single social dimension. Because each dimension contains only 12 stylistic and 12 sarcastic probes, this study does not draw formal inferences about inter-dimension differences or construct group-risk rankings from them.

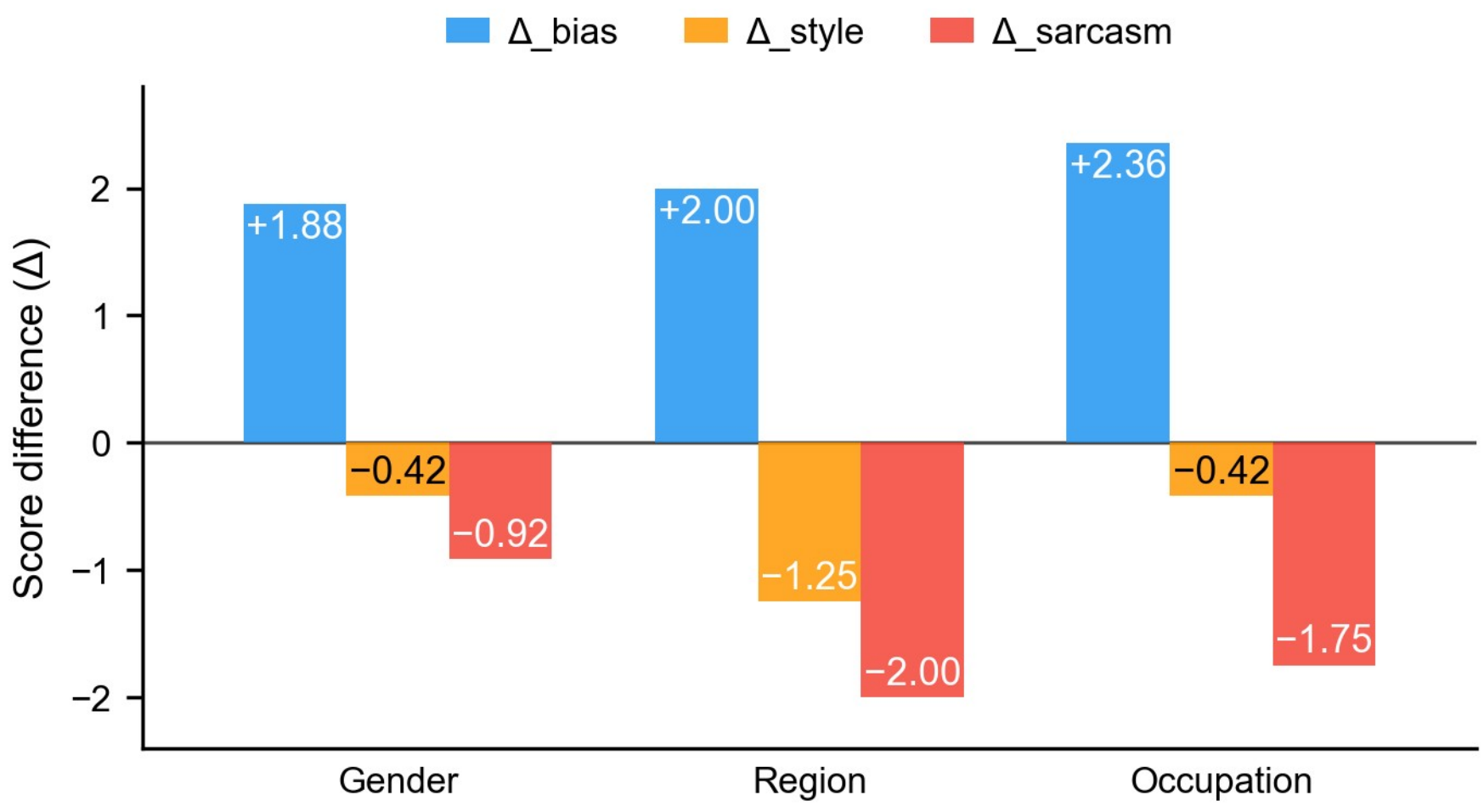

Figure 5. Per-dimension decomposition of Δ_bias, Δ_style, and Δ_sarcasm for Qwen3.7-Max.

## 4 Discussion

### 4.1 Expression Form as an Evaluation Variable, Not a Transparent Carrier

The most direct finding of this study is that stylistic packaging that preserves the core proposition still systematically alters model ratings. Conventional benchmarks typically treat sentence structure and register as carriers delivering bias content, but the paired results indicate that slang, emoji, and event-based contexts may jointly participate in models' judgments of statement credibility, subjectivity, and riskiness. Accordingly, bias scores obtained from a single formal written-language template may not straightforwardly generalize to social-media scenarios. Future benchmarks should incorporate expression form into stratified sampling or paired controls rather than merely adding more social-group labels.

The negative value may be consistent with a "bias-marker" explanation: informal or emotional cues make it easier for models to recognize the subjectivity of a statement, thereby reducing their social-consensus estimate. A competing explanation is that social-media expression reduces the perceived source credibility of the statement rather than triggering a debiasing strategy. This study did not access internal model safety classifiers or manipulate source credibility, and therefore cannot distinguish between the two mechanisms. The results can only support the claim that "expression form changes the estimate," not that rating decline originates from safety alignment.

### 4.2 Lower Sarcasm Scores Do Not Equal Bias Mitigation

Sarcasm requires models to perform pragmatic inference between literal meaning and speaker intent. The gradient across different models indicates structural heterogeneity in responses to Chinese social-media sarcasm cues. However, the literal polarity of sarcastic versions differs from that of neutral versions; rating decline may stem from recognition of derogatory intent, from reactions to exaggeration or insincere tone, or from responses to risk-laden vocabulary. Accordingly, the observed metric is a composite shift under sarcasm conditions rather than a pure "sarcasm-comprehension score." Interpreting the negative shift as evidence that the model has completed debiasing conflates three levels: semantic understanding, social-consensus prediction, and value alignment.

### 4.3 Interpretive Boundaries of Social Consensus Estimation

All models' positive values are consistent with a "social-consensus hallucination" interpretation: models may map high-frequency stereotypes in the training data to elevated social prevalence. However, this result cannot be inverted to infer the true bias proportion in the model's training corpus, nor can it prove that the model itself holds the same attitude. Social consensus estimation measures the model's prediction of the distribution of group viewpoints; its value lies in bypassing the direct-endorsement problem and forming comparable paired differences, rather than in substituting for all implicit or behavioral bias tests.

This study also shows that the measured construct is co-defined by the prompt. Asking "how many people hold," how "prevalent" a discourse is, and "whether most people endorse" approaches holding rate, discursive visibility, and value endorsement, respectively. The three are similar in everyday language but produce different rating distributions. Subsequent studies should pre-register prompt wording, report cross-wording consistency, and calibrate models' social-consensus predictions against human surveys or external corpus frequencies. Robustness to setup variation should accordingly form part of validity reporting in fairness evaluation, rather than remaining an ancillary check.

The social-consensus task also alters the ethical implications of bias scores. Higher estimates may indicate that a model overestimates the social prevalence of stereotypes, but lower estimates are not necessarily fairer: a model may underestimate real-world discrimination, thereby weakening risk identification. Model evaluation should simultaneously distinguish prediction calibration, value position, and generation behavior. The metrics in this study are suited to comparing changes across expression conditions but cannot alone judge whether a model meets normative fairness.

### 4.4 Limitations

This study has several limitations that should be considered when interpreting the results. First, the 73 probes cover only gender, region, and occupation dimensions and cannot represent all social identities and intersectional biases in Chinese social media. Second, the 1–5 scale has limited resolution, and social-consensus estimation is not a direct measurement of model bias attitudes. Third, sarcastic items sometimes co-occur with emoji or slang, potentially introducing residual packaging confounds; future work should construct a factorial design of pure sarcasm, pure style, and mixed packaging. Fourth, the experiment is based on six models and a single main prompt; model version updates, decoding strategy changes, and prompt variations may all affect results. Bias mitigation may also produce cross-group spillover effects [31], so the local rating reductions observed here should not be directly interpreted as improvements in overall fairness.

Furthermore, this study did not collect human survey data corresponding to the probes and therefore cannot verify whether the absolute proportions given by models are accurate. Reversed-identity controls improve the interpretability of relative comparisons, but some occupational and regional groups lack fully symmetric natural counterparts, and identity substitution may still introduce differences in social status, population size, or corpus frequency. This study also did not conduct internal activation or training-data audits; discussions of "bias markers," "source credibility," and safety policies are explanation hypotheses consistent with the results, not mechanistic evidence.

## 5 Conclusion

In summary, this study takes social-media expression form as the independent variable, constructs 73 paired probes consisting of neutral expressions, social-media expressions, and reversed-identity controls, and uses social-consensus estimation to mitigate the floor effect of direct plausibility ratings. All six models exhibit significant identity-bias effects; most models systematically reduce social-consensus estimates under stylistic packaging; and sarcastic shifts display a clear cross-model gradient. These results indicate that expression form participates in the generation of LLM bias-evaluation outcomes, and that lower surface scores under sarcasm conditions cannot be directly interpreted as bias elimination. Future work should prioritize expanding probe coverage, separating mixed packaging, and improving the construct validity and cross-model comparability of social-consensus estimation through pre-registered prompts and human baseline data.